\documentclass{fairmeta}

\newif\ifincludeappendix
\includeappendixtrue
\makeatletter
\@namedef{supplementlabel@alg:hypothesis-voi}{1}
\@namedef{supplementlabel@app:coevo}{H}
\@namedef{supplementlabel@app:fast-system-config}{E}
\@namedef{supplementlabel@app:libero-pro-examples}{A}
\@namedef{supplementlabel@app:memory}{C}
\@namedef{supplementlabel@app:ppo-training}{F}
\@namedef{supplementlabel@app:real-robot-training}{B}
\@namedef{supplementlabel@app:shared-backbones}{I}
\@namedef{supplementlabel@app:voi}{D}
\makeatother
\newcommand{\supplementref}[1]{%
  \ifincludeappendix\ref{#1}\else\csname supplementlabel@#1\endcsname\fi}
\newcommand{\supplementautoref}[1]{%
  \ifincludeappendix\autoref{#1}\else Appendix~\supplementref{#1}\fi}

\usepackage{amsmath,amsfonts,bm}

\def\eqref#1{equation~\ref{#1}}

\def\1{\bm{1}}

\DeclareMathAlphabet{\mathsfit}{\encodingdefault}{\sfdefault}{m}{sl}
\SetMathAlphabet{\mathsfit}{bold}{\encodingdefault}{\sfdefault}{bx}{n}

\usepackage{hyperref}
\usepackage{url}
\usepackage{booktabs}
\usepackage{graphicx}
\usepackage{xcolor}
\usepackage{colortbl}
\usepackage{wrapfig}
\usepackage{float}
\usepackage{algorithm}
\usepackage{algpseudocode}

\newcommand{\myparagraph}[1]{\vspace{2pt}\textbf{#1}}

\definecolor{positiontint}{RGB}{225,235,244}
\definecolor{tasktint}{RGB}{244,236,221}
\definecolor{unseentint}{RGB}{231,241,231}
\definecolor{ourslavender}{RGB}{239,233,249}
\newcommand{\highlightrowstrut}{\rule[-0.8ex]{0pt}{3.6ex}}

\makeatletter
\newcommand{\solidrowcolor}[3]{\noalign{\begingroup
  \color{#1}%
  \dimen@=#2\relax
  \ifdim\ht\@arstrutbox>\dimen@ \dimen@=\ht\@arstrutbox\fi
  \@tempdima=#3\relax
  \ifdim\dp\@arstrutbox>\@tempdima \@tempdima=\dp\@arstrutbox\fi
  \advance\dimen@ by\@tempdima
  \hrule height\dimen@\kern-\dimen@
\endgroup}}
\makeatother

\title{EmbodiedRSI: Active Continual Robot Learning Through Hypothesis-Guided Co-Evolution}

\author[1]{Python Song}
\author[2,\dagger]{Zhixuan Liang}
\author[2]{Kelsey Fu}
\author[2]{Mengdi Wang}
\author[1]{Junfeng Yang}
\author[2,1,\dagger]{Shilong Liu}

\affiliation[1]{Columbia University}
\affiliation[2]{Princeton University}
\contribution[\dagger]{Corresponding Authors}

\abstract{
Robot foundation models provide strong visuomotor control, yet their performance can degrade when object positions or task instructions change. Further improvements often require post-training on substantial robot data, which can be costly to collect through methods such as teleoperation. Agentic harnesses can adapt around the model, but current self-evolving harnesses use robot trials inefficiently when deciding which code and skill changes to pursue. 
We introduce \textbf{EmbodiedRSI}, a self-evolving agentic harness that autonomously decides where to explore next and turns the resulting physical interaction into improved code and skills.
EmbodiedRSI realizes this through a Fast-Slow Dual-System Architecture, in which competing code and skill hypotheses are maintained in a Hypothesis Graph. Value-of-Information Experiment Selection chooses physical experiments that can distinguish these hypotheses. Their outcomes guide Code-Skill Co-Evolution. The Slow System builds Hierarchical Memory, and Reward-Grounded Memory Learning selects effective memory according to their value for later Fast-System improvement.
On RoboCasa365, EmbodiedRSI reaches \textbf{77.0\%} overall success and \textbf{71.3\%} on Composite-Unseen, compared with \textbf{40.1\%} for the best baseline. EmbodiedRSI also reaches \textbf{86.8\%} overall success on LIBERO-Pro. Beyond benchmark performance, EmbodiedRSI transfers zero-shot to real-world robot, achieving \textbf{71.3\%} overall success across multiple challenging tasks.
}

\titlefooter{%
  {\footnotesize\textbf{Website:} \url{https://geeksongs.github.io/agentic_robotics/}}\hfill
  \raisebox{-0.5\height}{\includegraphics[height=0.630cm]{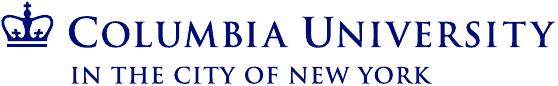}}%
  \hspace{0.3cm}%
  \raisebox{-0.5\height}{\includegraphics[height=0.688cm]{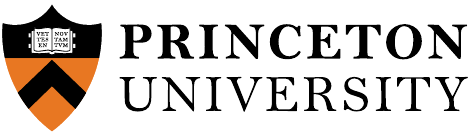}}%
}

\begin{document}
\maketitle

\section{Introduction}

\begin{wrapfigure}[22]{r}{0.42\textwidth}
\vspace{-\intextsep}\vspace{-12pt}
\centering
\includegraphics[width=0.98\linewidth]{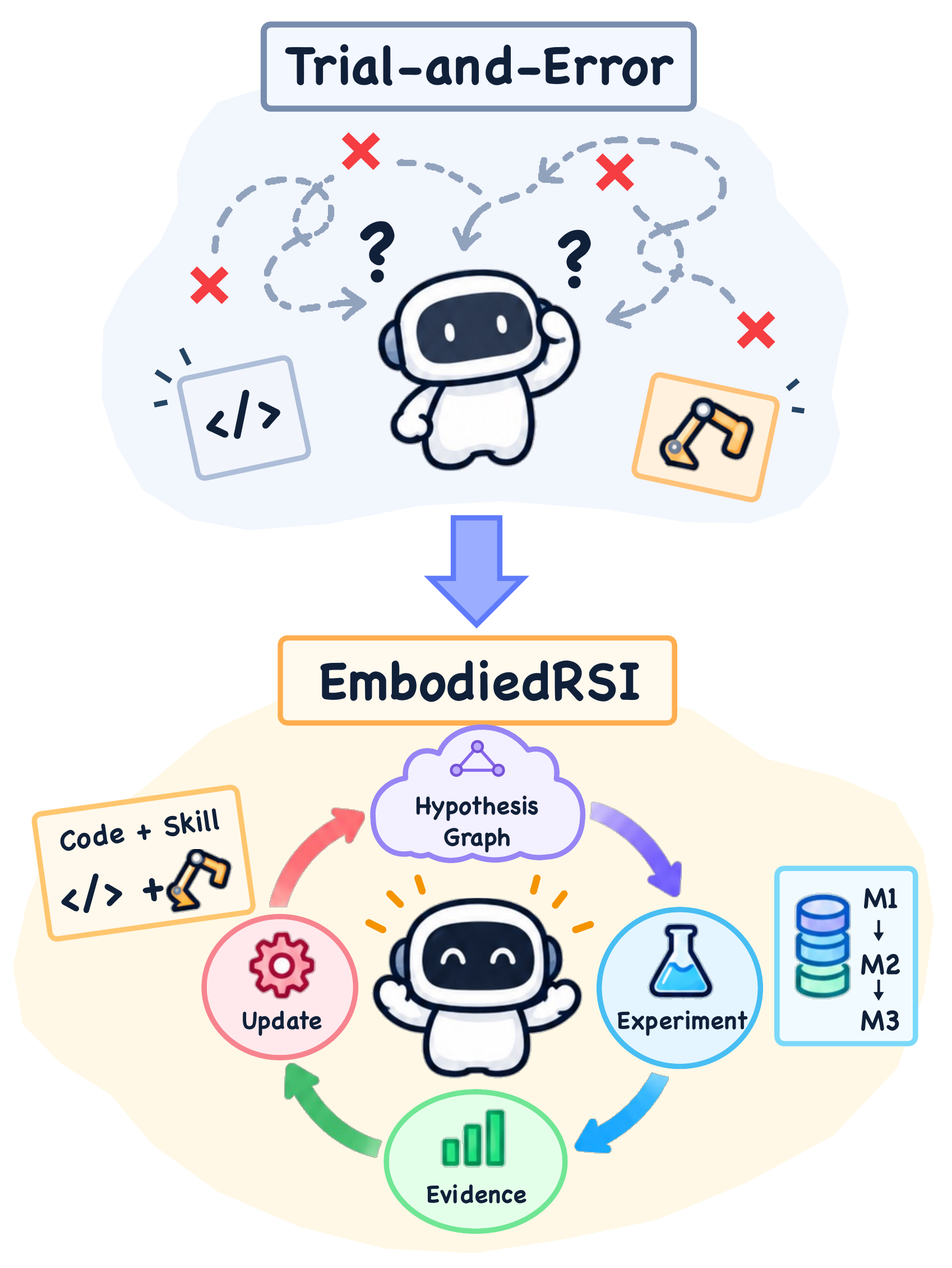}
\vspace{-12pt}
\caption{\textbf{The EmbodiedRSI Self-Evolution Loop.} Physical outcomes guide code and skill updates while the robot foundation model remains frozen.}
\label{fig:sticker-loop}
\vspace{-\intextsep}
\end{wrapfigure}

Robot foundation models have substantially advanced general-purpose robot learning, enabling transferable behavior across diverse tasks and environments. However, their performance can degrade when deployment conditions depart from prior experience, such as changes in object positions or task instructions \citep{zhou2025liberopro}. Adapting the underlying model typically requires additional robot data collection and post-training, while gathering sufficient new interaction data for finetuning remains costly. Agentic robot harnesses provide an alternative by adapting robot behavior at the execution layer through a small number of trial executions \citep{liang2023code,zhang2026harnessvla,fu2026capx,ding2026zetta,lu2026aspire}. These harnesses use high-level reasoning to organize robot execution, and update themselves based on the outcomes of a few attempts. Some synthesize and revise executable code after observing execution failures, while others coordinate pretrained robot policies through planning, memory, and feedback. More recent systems further enable the harness to evolve from interaction, allowing skills to improve as experience accumulates.

Despite these advances, current self-evolving robotics harnesses still learn inefficiently from physical interaction. As the space of executable programs, behavioral skills, and their compositions grows, iterative proposal-and-execution loops can repeatedly pursue closely related improvement directions without explicitly deciding \textit{which physical experiment would provide the most useful evidence for the next update}.
Physical trials remain costly, yet existing harnesses treat them as an undifferentiated resource to consume rather than a scarce resource to actively allocate.
This inefficiency is compounded by two further gaps.
Code and skills are also commonly evolved through separate processes, even though long-horizon manipulation often depends on their interaction, and memory is typically accumulated for retrieval and reuse rather than learned according to whether it improves later self-evolution.
As a result, existing harnesses lack a principled way to direct physical exploration toward the most informative experiments, while still coordinating code-skill improvement and retaining experience useful for future adaptation. We therefore ask how an agentic robotics harness can decide what experiment to run next, so as to efficiently transform physical interaction into continual Code-Skill Co-Evolution and reusable knowledge for adaptation to unseen scenes and long-horizon tasks.

We therefore present \textbf{EmbodiedRSI}, an agentic harness in which each physical episode informs which code or skill hypothesis to pursue next. A Fast System proposes and executes candidate experiments, each stating which hypotheses its outcome can distinguish, and revises its confidence in response to every physical outcome, while a Hypothesis Graph preserves the refinement history to inform later decisions. Within this loop, Code-Skill Co-Evolution compares each candidate pair with its individual components from the same initial states, showing that code and skills reinforce each other beyond what either achieves alone and using this evidence to choose the next physical experiment.
Hierarchical Memory further distills the refinement history into reusable cognition, managed according to downstream value to the Fast System (measured by later task improvement, exploration-uncertainty reduction, and physical-experiment efficiency) through reinforcement learning rather than fixed heuristics.

Our experiments establish that a self-evolving agentic harness around a frozen robot foundation model turns physical interaction directly into reusable code and skill improvements that generalize to unseen scenes and tasks, while preserving the underlying model's local control competence. 

Our main contributions share a common goal, enabling the harness to autonomously decide where to explore so that it learns from physical interaction more efficiently. \textbf{(i)} A \textbf{Fast-Slow Dual-System Architecture} realizes this principle directly, letting the Fast System select the next physical experiment using current hypothesis weights and cost and rapidly iterate code and skill updates, while the Slow System builds Hierarchical Memory that abstracts past experience to inform this choice. \textbf{(ii)} \textbf{Code-Skill Co-Evolution} lets code and skill updates reinforce each other rather than evolve separately, comparing each candidate pair against its individual components from the same initial states, with a Hypothesis Graph keeping these comparisons traceable to the evidence behind them. \textbf{(iii)} \textbf{Reward-Grounded Memory Learning} completes the loop, training the selection of memory actions according to downstream value to the Fast System. \textbf{(iv)} On RoboCasa365, EmbodiedRSI reaches state-of-the-art \textbf{77.0\%} overall success, including \textbf{71.3\%} on Composite-Unseen, and \textbf{86.8\%} overall success on LIBERO-Pro. Beyond benchmark performance, EmbodiedRSI evolves its agent pipeline entirely in simulation and transfers it zero-shot to real-world experiments, achieving \textbf{71.3\%} overall success across multiple challenging tasks.

\section{Related Work}
\label{sec:related}

\myparagraph{Agentic Execution Around Frozen Robot Foundation Models.}
Generalist vision-language-action (VLA) policies improve transfer through scale and vision-language pretraining \citep{brohan2022rt1,brohan2023rt2,kim2024openvla}. Frozen-VLA agents add planning, memory, and tools around the policy. Harness VLA uses a frozen VLA as a contact-rich primitive \citep{zhang2026harnessvla}, while Zetta evolves runtime critics, response skills, and tools at multiple time scales \citep{ding2026zetta}. EmbodiedRSI links physical attempts, candidate code or skill updates, and outcomes in a Hypothesis Graph before Code-Skill Co-Evolution constructs an update.

\myparagraph{Skill Libraries and Code-Mediated Improvement.}
Code as Policies writes robot-control programs from language and perception APIs \citep{liang2023code}. CaP-X and Playful Agentic Robot Learning build code-skill libraries from successful executions \citep{fu2026capx,zhang2026playful}, while Voyager accumulates executable skills from environment feedback \citep{wang2023voyager}. Other self-evolution methods revise skills or optimize an external skill and harness layer \citep{ju2026embodiskill,wang2026shaper}, and Eureka evolves reward programs \citep{ma2023eureka}. EmbodiedRSI represents competing explanations in a Hypothesis Graph, uses Value-of-Information Experiment Selection to select physical experiments, and retains code and skill updates after trials from the same initial states.

\myparagraph{Embodied Memory and Continual Learning.}
Embodied-RAG develops hierarchical memory for embodied retrieval and generation \citep{xie2024embodiedrag}, and MemSkill learns memory-writing skills \citep{zhang2026memskill}. EmbodiedRSI uses physical transitions to build reusable memory and learns from later task outcomes which memory actions support self-improvement.

\section{Method}
\label{sec:method}

\begin{figure}[tb]
\centering
\includegraphics[width=\textwidth]{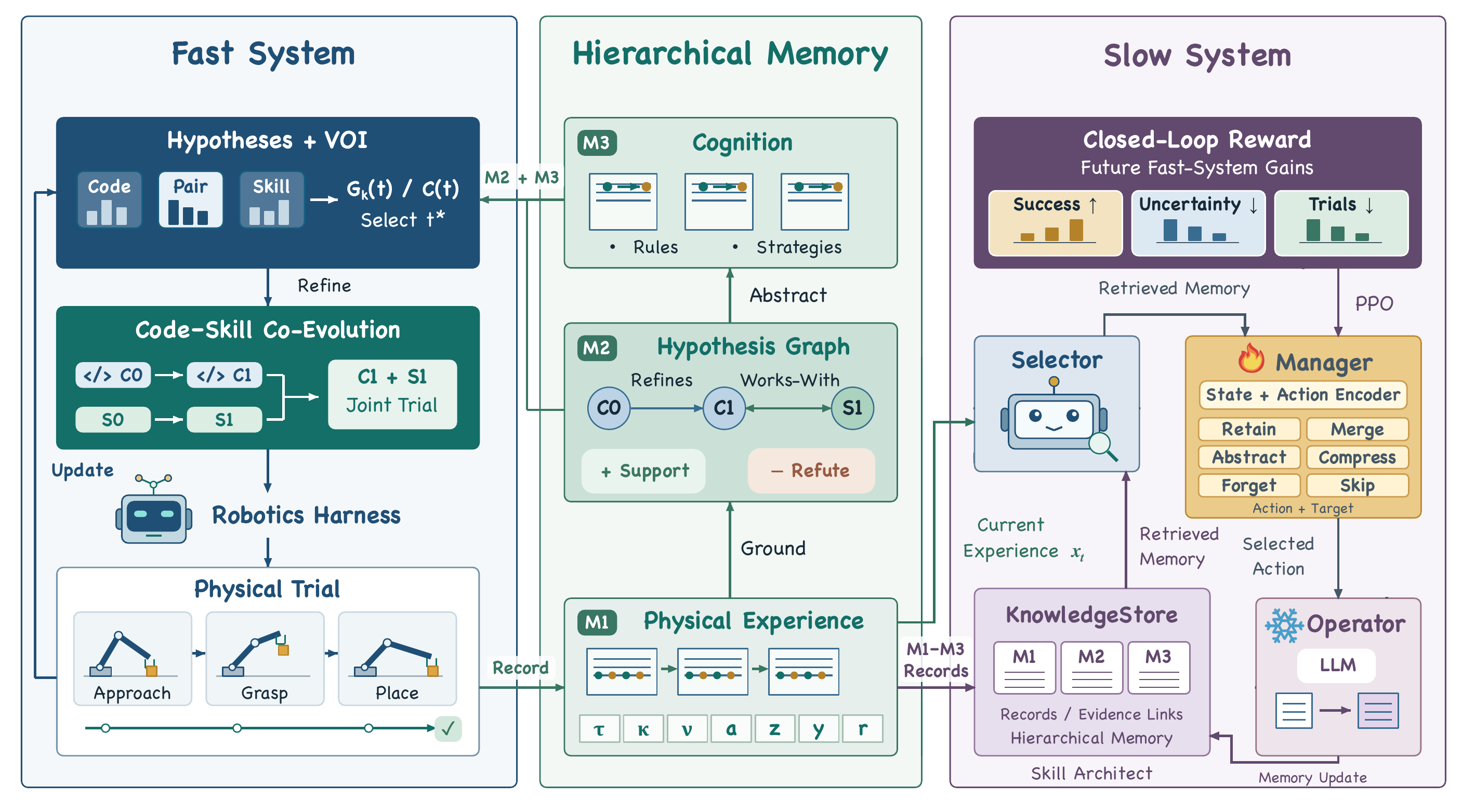}
\vspace{-20pt}
\caption{\textbf{Overview of EmbodiedRSI.} The Fast System uses a Hypothesis Graph to choose each physical experiment. Each result guides a joint code-skill update. The Slow System retains memories that improve later adaptation.}
\label{fig:embodied-rsi-overview}
\end{figure}

\subsection{Problem Formulation}
We formulate EmbodiedRSI as an agentic robotics harness that continually evolves through physical interaction. After $k$ physical episodes, $\mathcal{A}_k$ denotes the current harness, including its executable code and skills, hypothesis set $\mathcal{H}_k$, and accumulated memory $\mathcal{M}_k$.

At each episode, EmbodiedRSI determines the next physical experiment from the candidate set $\mathcal{T}_k$ based on its current hypotheses and memory:
\begin{equation}
    t_k^{*} = f(\mathcal{T}_k,\mathcal{H}_k,\mathcal{M}_k),
\end{equation}
where $f$ represents the experiment-selection process and $t_k^{*}$ is the physical experiment performed at episode $k$.
Executing $t_k^{*}$ produces physical evidence $e_k$, including the execution trajectory and task outcome. The harness then evolves according to
\begin{equation}
    \mathcal{A}_{k+1} = g(\mathcal{A}_k,t_k^{*},e_k),
\end{equation}
where $g$ represents the evolution process that incorporates new physical evidence into the harness.

Given a budget of $B$ physical episodes, this process yields $\mathcal{A}_B$. For $N$ evaluation episodes with binary task outcomes $b_i(\mathcal{A}_B)\in\{0,1\}$, the task success rate $R_{\mathrm{task}}(\mathcal{A}_B)$ is the fraction of successful evaluations. Our objective is to maximize this task success rate:
\begin{equation}
    R_{\mathrm{task}}(\mathcal{A}_B)=\frac{1}{N}\sum_{i=1}^{N}b_i(\mathcal{A}_B),\qquad
    \max_{\mathcal{A}_B} R_{\mathrm{task}}(\mathcal{A}_B).
\end{equation}

\subsection{Fast System: Rapid Code-Skill Co-Evolution}
\label{sec:hgraph}

\myparagraph{Hypothesis Graph and Confidence.} Each selected experiment produces evidence about possible code and skill changes. To keep that evidence tied to hypotheses as they evolve, the Hypothesis Graph maintains code and skill hypotheses as nodes, with a directed $\textsc{refines}$ edge linking each hypothesis to its revision. The Fast System uses each trajectory and outcome to generate new code or skill hypotheses and to refine existing hypotheses when the evidence supports a more specific version. To track how strongly physical experiments support each hypothesis $h_i$, all hypotheses begin with the same $\operatorname{Beta}(\alpha,\beta)$ prior on their support rate~\citep{gelman2013bayesian}, where $\alpha$ and $\beta$ are prior support and refutation counts. Let $n_{\mathrm{sup},i}$ count physical experiments supporting $h_i$ and $n_{\mathrm{ref},i}$ count outcomes inconsistent with $h_i$. Matched trials update each tested hypothesis from its success-rate difference, holding initial states and all other calls and their order constant. The posterior mean confidence of $h_i$ is:
\begin{equation}
    c_i = \frac{n_{\mathrm{sup},i}+\alpha}{n_{\mathrm{sup},i}+n_{\mathrm{ref},i}+\alpha+\beta}.
\end{equation}
The mean confidence captures support for each hypothesis; choosing the next experiment also requires knowing where that support remains uncertain. Let $D_k$ be the matched outcomes observed after $k$ physical episodes and $q_i=P(\theta_i>1/2\mid D_k)$, where $\theta_i$ is the support rate of $h_i$. Multiple hypotheses can jointly contribute to task success. We measure uncertainty over their support as:
\begin{equation}
    \mathcal{H}_{\mathrm{exp}}(D_k)=\sum_{h_i\in\mathcal{H}}h_2(q_i),
    \quad h_2(q)=-q\log_2q-(1-q)\log_2(1-q).
\end{equation}
A matched trial provides initial evidence for each new hypothesis before its subsequent experiments are selected adaptively.

\myparagraph{Value-of-Information Experiment Selection.} Physical outcomes update the confidence of each hypothesis, revealing where evidence is still needed. To decide which uncertainty to address within its limited interaction budget, the Fast System constructs experiments $t\in\mathcal{T}$ that evaluate code and skill hypotheses through matched trials. Each experiment declares the subset $S_t\subseteq\mathcal{H}$ whose confidence its outcome can update. Given the evidence $D_k$, Value-of-Information Experiment Selection measures the expected reduction in exploration uncertainty from the possible outcomes $e$ of experiment $t$:
\begin{equation}
    G_k(t)=\mathcal{H}_{\mathrm{exp}}(D_k)
    -\mathbb{E}_{e\sim P(e\mid D_k,t)}
    \bigl[\mathcal{H}_{\mathrm{exp}}(D_k\cup\{e\})\bigr].
\end{equation}
Appendix~\supplementref{app:voi} gives the outcome probabilities. Let $T_t$, $N_{\mathrm{step},t}$, $C_{\mathrm{GPU},t}$, and $C_{\mathrm{reset},t}$ denote elapsed time, simulator steps, GPU cost, and reset cost, with weights $w_\tau$, $w_s$, $w_g$, and $w_r$. The physical-experiment cost is $C(t)=w_\tau T_t+w_sN_{\mathrm{step},t}+w_gC_{\mathrm{GPU},t}+w_rC_{\mathrm{reset},t}$, the selection score is $\operatorname{VOI}_k(t)=G_k(t)/C(t)$, and the next experiment is $t_k^*=\arg\max_{t\in\mathcal{T}}\operatorname{VOI}_k(t)$.
After executing $t_k^*$, the Fast System updates the evaluated hypotheses from matched outcomes. The resulting posteriors determine both exploration uncertainty and the next experiment ranking. Algorithm~\supplementref{alg:hypothesis-voi} in Appendix~\supplementref{app:voi} gives the complete procedure. The experimental configuration is provided in \supplementautoref{app:fast-system-config}. With thresholds $\epsilon_H$ and $\epsilon_{\mathrm{VOI}}$, broad exploration ends when $\mathcal{H}_{\mathrm{exp}}<\epsilon_H$, $\max_t\operatorname{VOI}_k(t)<\epsilon_{\mathrm{VOI}}$, the rollout budget is exhausted, or every hypothesis reaches the evaluation cap $n_{\max}$.

\begin{figure}[t]
\centering
\includegraphics[width=\textwidth]{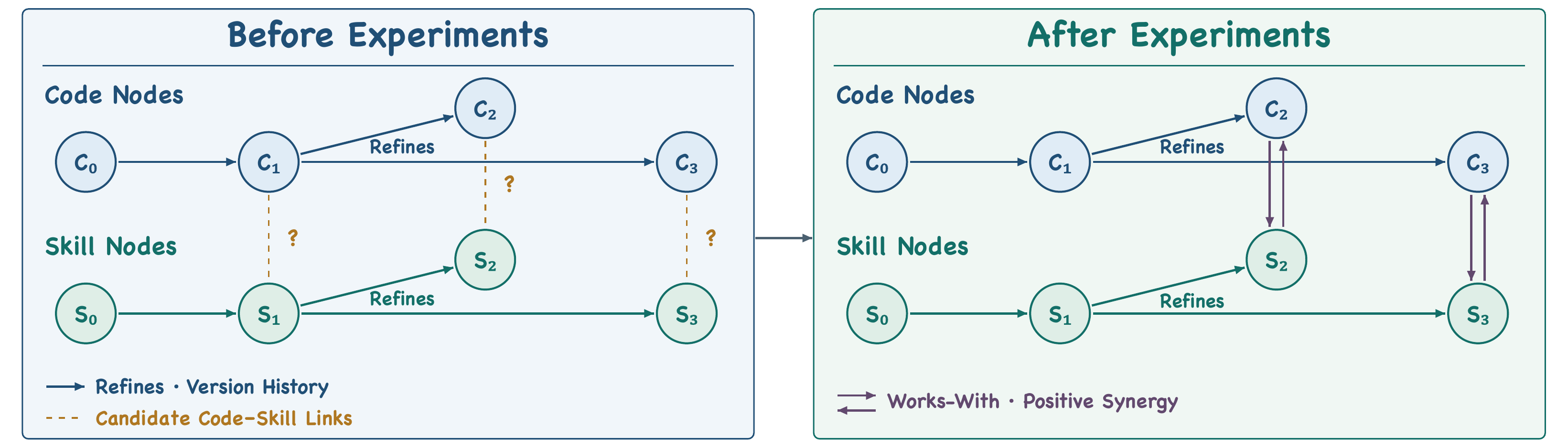}
\vspace{-20pt}
\caption{\textbf{Hypothesis Graph Updates from Physical Trials.} \textbf{Before Experiments (left).} Directed \textsc{refines} edges show how code and skill hypotheses have evolved, while dashed links mark code-skill relations awaiting evidence. \textbf{After Experiments (right).} Solid bidirectional \textsc{works-with} edges identify relations supported by measured positive joint gain.}
\label{fig:hypothesis-graph-update}
\vspace{-2pt}
\end{figure}

\myparagraph{Code-Skill Co-Evolution.} The selected experiment provides evidence for deciding what to update. An improvement may come from code, skills, or their interaction. Code-Skill Co-Evolution separates code, skill, and joint effects by evaluating code hypothesis $C_i$, skill hypothesis $S_j$, and their joint execution from the same initial states. Let $R(C_i,S_j)$, $R(C_i)$, and $R(S_j)$ denote the measured task success rates of joint, code-only, and skill-only execution, respectively. Their measured joint gain is:
\begin{equation}
    w_{ij}=R(C_i,S_j)-\max\{R(C_i),R(S_j)\}.
\end{equation}
Positive joint gain indicates a higher success rate for joint execution in the matched trials. The Slow System records this relation as a bidirectional $\textsc{works-with}$ edge in the Hypothesis Graph, prioritizing the strongest relations while directed revision edges retain the refinement history. The Fast System evaluates new and revised hypotheses in subsequent experiments and retains executable improvements for the next round. Each outcome enters Hierarchical Memory and changes what the Fast System will explore next. Figure~\ref{fig:hypothesis-graph-update} shows how physical trials turn candidate code-skill connections into retained $\textsc{works-with}$ relations.

\subsection{Slow System: Hierarchical Memory and Reward-Grounded Memory Learning}
\label{sec:memlearn}

\myparagraph{Hierarchical Memory.}
\label{sec:memory}
Each experiment produces evidence that later decisions need at different levels of detail. Hierarchical Memory carries this evidence forward at three levels: M1 records physical execution, M2 maintains code and skill hypotheses in the Hypothesis Graph, and M3 retains reusable cognition distilled from repeated evidence.

\begin{samepage}
For an episode, let $\tau$ be the robot execution trajectory, $\kappa$ the invoked tools, $\nu$ the time steps at which the VLA is invoked, $a$ the executed actions, $z$ the telemetry, $y$ the physical outcome, and $r$ the task reward. The M1 experience record is:
\begin{equation}
    \mathcal{M}_{1} = \{(\tau, \kappa, \nu, a, z, y, r)\}.
\end{equation}
\end{samepage}
M1 automatically records every episode and supplies matched evidence for M2 count and relation updates under the Fast-System rules. Across episodes, the Slow System abstracts recurring M1 and M2 evidence into M3. The hierarchy is:
\begin{equation}
    \mathcal{M}_{1} \rightarrow \mathcal{M}_{2} \rightarrow \mathcal{M}_{3}.
\end{equation}
Before the next physical episode, the Fast System retrieves M2 hypotheses and M3 cognition to guide its candidate experiments.

\myparagraph{Reward-Grounded Memory Learning.} For the memory hierarchy to guide later experiments, the Slow System learns how to transform incoming evidence into reusable knowledge. Reward-Grounded Memory Learning bases that decision on later Fast-System improvement. Using the current experience and memory, the Manager selects an action and target records; the frozen operator executes it to update knowledge retrieved in the next round. For current experience $x_t$, the Slow System retrieves $N_{\mathrm{ret}}$ relevant records $\{m_{t,\ell}\}_{\ell=1}^{N_{\mathrm{ret}}}$ from M1, M2, and M3, including hypothesis confidence and evidence links. We encode experience, retrieved records, and action descriptions with the frozen Qwen3-Embedding-0.6B encoder $e(\cdot)$~\citep{zhang2025qwen3embedding}. The trainable context function $f_{\theta}^{\mathrm{ctx}}$ aggregates these embeddings into Slow-System state $h_t$.
Each candidate in $\mathcal{S}=\{s_1,\ldots,s_K\}$ specifies an applicable action and target record or evidence group. The action space comprises Retain, Merge, Abstract, Compress, Forget, and Skip. Let $u_i$ represent candidate $s_i$, and let $z_{t,i}$ measure compatibility between $h_t$ and $u_i$. The Slow-System policy is $\pi_{\theta}(i\mid h_t)=\operatorname{softmax}(z_t)_i$.
Here $z_t=(z_{t,1},\ldots,z_{t,K})$ collects the compatibility scores and $[\cdot,\cdot]$ denotes concatenation. $\operatorname{desc}(s_i)$ describes the action and target of candidate $s_i$. The trainable skill function $f_{\theta}^{\mathrm{skill}}$ maps each encoded description to $u_i$, and the trainable scoring function $f_{\theta}^{\mathrm{score}}$ maps $[h_t,u_i]$ to $z_{t,i}$. The operator uses frozen Qwen3.5-122B-A10B~\citep{qwenteam2026qwen35} to update the retrieval index or M3 according to the selected candidate, preserving raw M1 records and M2 evidence updates. Appendix~\supplementref{app:memory} specifies the actions.

\myparagraph{Closed-Loop Reward and RL Training.} The value of each memory action becomes visible through its effect on later Fast-System experiments. At memory decision step $t$, let $R_t$ estimate $R_{\mathrm{task}}$ for the current harness, $H_t$ denote the exploration uncertainty $\mathcal{H}_{\mathrm{exp}}$, and $C_{t+1}$ be the realized cost of the next physical experiment under $C(\cdot)$. We compare success rates under the same evaluation protocol and compute entropy changes over the same hypothesis set. With reward weights $\lambda_1$, $\lambda_2$, and $\lambda_3$, the reward following the next experiment is:
\begin{equation}
    r_t^{\mathrm{mem}}=\lambda_1(R_{t+1}-R_t)+\lambda_2(H_t-H_{t+1})-\lambda_3 C_{t+1}.
\end{equation}
For a horizon of $L$ memory decisions and discount factor $\gamma\in(0,1]$, the discounted return is $J_t^{\mathrm{mem}}=\sum_{j=0}^{L-1}\gamma^j r_{t+j}^{\mathrm{mem}}$.
The Slow-System policy is trained with PPO~\citep{schulman2017proximal} to maximize expected discounted return. Training settings are given in \supplementautoref{app:ppo-training}. An architect can refine action instructions from recurring experience. Thus each episode can change both the hypotheses considered by the Fast System and the memories available when the next experiment is selected.

\section{Experiments}
\label{sec:experiments}

Our experiments address three questions about generalization, zero-shot sim-to-real transfer, and learning efficiency. \textbf{Q1}: How does EmbodiedRSI perform on RoboCasa365 and LIBERO-Pro, especially under cross-task, cross-scene, and out-of-distribution generalization? \textbf{Q2}: Can EmbodiedRSI's agentic harness transfer zero-shot from simulation to a real robot? \textbf{Q3}: Which components of EmbodiedRSI improve success and learning efficiency?

\subsection{Experimental Setup}
\label{sec:setup}

\myparagraph{Benchmark and Backbone.} We evaluate on two simulated benchmarks. We select RoboCasa365 to evaluate generalization across household scenes and task compositions, and LIBERO-Pro to examine changes in task goals and object layouts. RoboCasa365 \citep{nasiriany2026robocasa365} extends RoboCasa's diverse household scenes and compositional manipulation tasks \citep{nasiriany2024robocasa} to large-scale mobile manipulation, with Atomic-Seen, Composite-Seen, and Composite-Unseen task suites. LIBERO-Pro \citep{zhou2025liberopro} extends LIBERO \citep{liu2023libero} with controlled perturbations. We evaluate instruction-redirection (T) and position-swap (S) perturbations. Backbone settings are given in \supplementautoref{app:shared-backbones}. On both benchmarks, we evaluate each task with 10 random seeds and 10 trials per seed. Evolution uses 10 separate seeds, while all methods are evaluated on another 10 mutually disjoint seeds.

\myparagraph{Baselines.} We compare against end-to-end frozen-VLA policies (OpenVLA~\citep{kim2024openvla}, $\pi_0$~\citep{black2025pi0}, $\pi_{0.5}$~\citep{physicalintelligence2025pi05}, $\pi_{\text{RLinf}}$~\citep{yu2025rlinf}, RLDX-1~\citep{kim2026rldx1}, NORA~\citep{hung2025nora}, MolmoAct~\citep{lee2025molmoact}, X-VLA~\citep{zheng2025xvla}, AtomVLA~\citep{sun2026atomvla}), a world-model baseline (WorldDreamer~\citep{wang2024worlddreamer}), a leaderboard entry (Xiaomi-Robotics-1), and code-as-policy and harness agents built on a frozen VLA (CaP-X~\citep{fu2026capx}, RATS~\citep{zhang2026playful}, Harness VLA~\citep{zhang2026harnessvla}, and Zetta~\citep{ding2026zetta}).

\myparagraph{Metrics.} On \textbf{RoboCasa365}, we report success on the Atomic-Seen, Composite-Seen, and Composite-Unseen (out-of-distribution) suites together with the overall success rate. On \textbf{LIBERO-Pro}, we report success on each task-family/perturbation cell together with the overall success rate.

\begin{figure}[!t]
\centering
\begin{minipage}[t]{0.32\textwidth}
    \centering
    \begin{minipage}[t]{0.48\linewidth}
        \centering
        {\setlength{\fboxsep}{1.5pt}\colorbox{positiontint}{\makebox[\dimexpr\linewidth-2\fboxsep\relax][c]{\tiny\bfseries Counter to Cabinet}}}
        \par\vspace{1pt}
        \includegraphics[width=\linewidth,trim=240bp 0bp 16bp 0bp,clip]{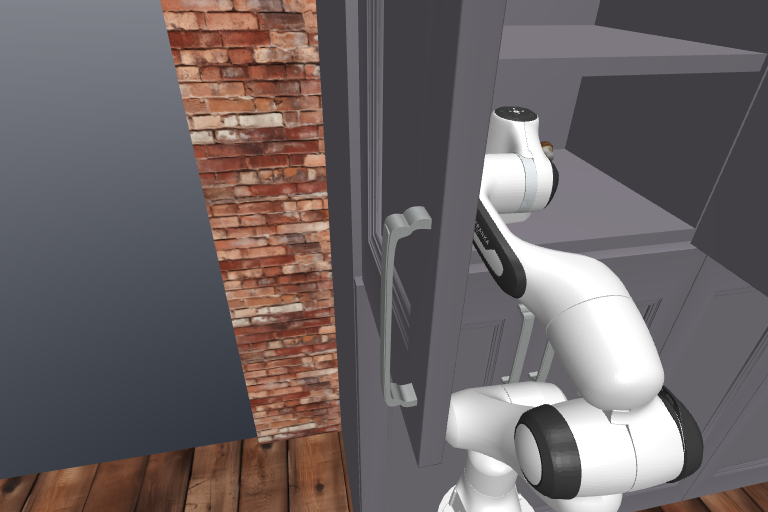}
    \end{minipage}\hfill
    \begin{minipage}[t]{0.48\linewidth}
        \centering
        {\setlength{\fboxsep}{1.5pt}\colorbox{positiontint}{\makebox[\dimexpr\linewidth-2\fboxsep\relax][c]{\tiny\bfseries Turn On Microwave}}}
        \par\vspace{1pt}
        \includegraphics[width=\linewidth,trim=240bp 0bp 16bp 0bp,clip]{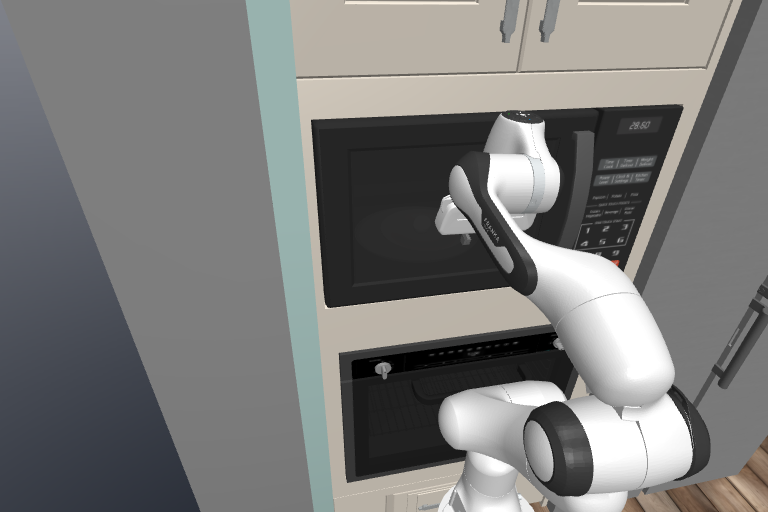}
    \end{minipage}
    \par\smallskip\footnotesize\textbf{(a)} Atomic-Seen
\end{minipage}\hfill
\begin{minipage}[t]{0.32\textwidth}
    \centering
    \begin{minipage}[t]{0.48\linewidth}
        \centering
        {\setlength{\fboxsep}{1.5pt}\colorbox{tasktint}{\makebox[\dimexpr\linewidth-2\fboxsep\relax][c]{\tiny\bfseries Load Dishwasher}}}
        \par\vspace{1pt}
        \includegraphics[width=\linewidth,trim=240bp 0bp 16bp 0bp,clip]{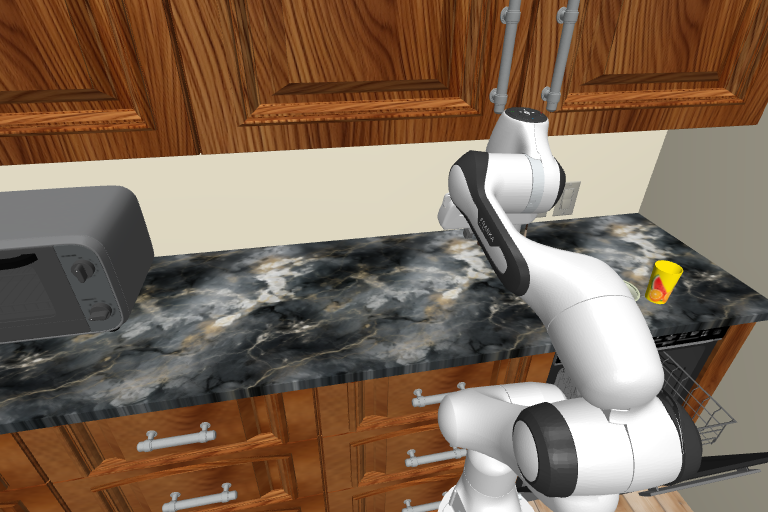}
    \end{minipage}\hfill
    \begin{minipage}[t]{0.48\linewidth}
        \centering
        {\setlength{\fboxsep}{1.5pt}\colorbox{tasktint}{\makebox[\dimexpr\linewidth-2\fboxsep\relax][c]{\tiny\bfseries Prepare Coffee}}}
        \par\vspace{1pt}
        \includegraphics[width=\linewidth,trim=240bp 0bp 16bp 0bp,clip]{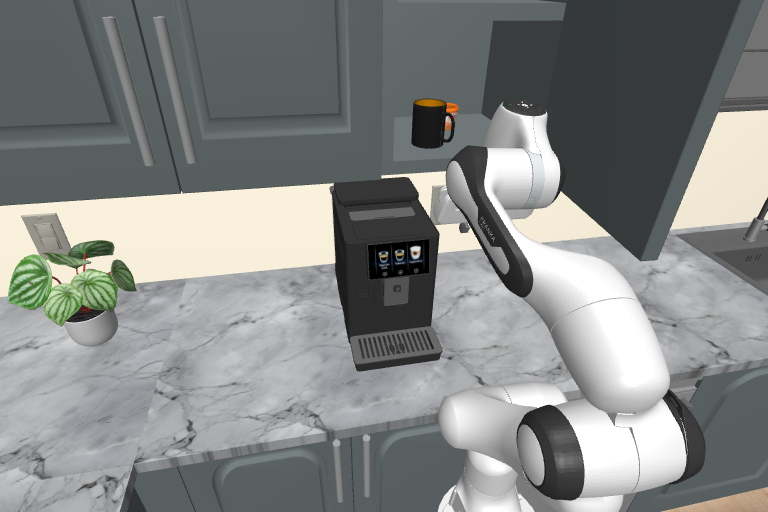}
    \end{minipage}
    \par\smallskip\footnotesize\textbf{(b)} Composite-Seen
\end{minipage}\hfill
\begin{minipage}[t]{0.32\textwidth}
    \centering
    \begin{minipage}[t]{0.48\linewidth}
        \centering
        {\setlength{\fboxsep}{1.5pt}\colorbox{unseentint}{\makebox[\dimexpr\linewidth-2\fboxsep\relax][c]{\tiny\bfseries Arrange Tea}}}
        \par\vspace{1pt}
        \includegraphics[width=\linewidth,trim=240bp 0bp 16bp 0bp,clip]{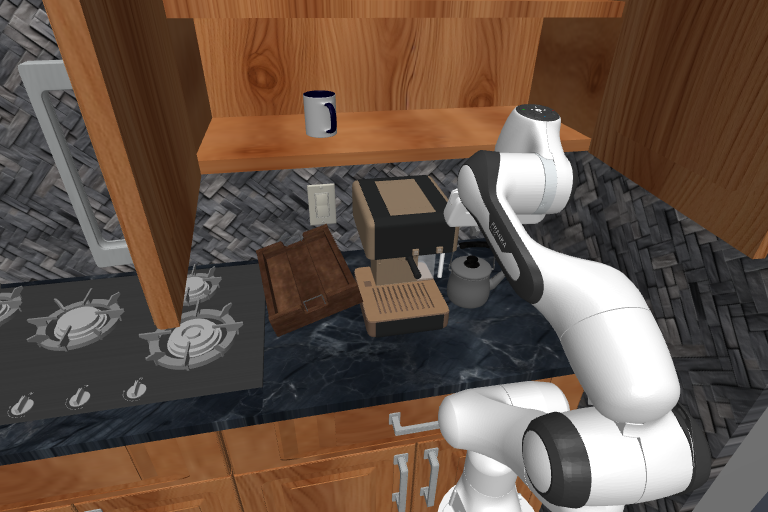}
    \end{minipage}\hfill
    \begin{minipage}[t]{0.48\linewidth}
        \centering
        {\setlength{\fboxsep}{1.5pt}\colorbox{unseentint}{\makebox[\dimexpr\linewidth-2\fboxsep\relax][c]{\tiny\bfseries Make Ice Lemonade}}}
        \par\vspace{1pt}
        \includegraphics[width=\linewidth,trim=240bp 0bp 16bp 0bp,clip]{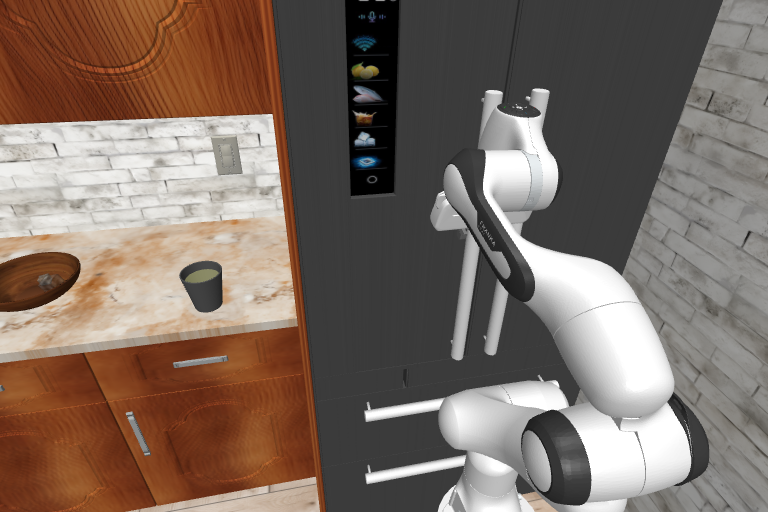}
    \end{minipage}
    \par\smallskip\footnotesize\textbf{(c)} Composite-Unseen
\end{minipage}
\vspace{-5pt}
\caption{\textbf{RoboCasa365 Task Suites.} \textbf{(a) Atomic-Seen} contains individual manipulation tasks observed during pretraining. \textbf{(b) Composite-Seen} combines multiple skills into tasks represented in pretraining. \textbf{(c) Composite-Unseen} combines skills into tasks absent from pretraining.}
\label{fig:robocasa365-tasks}
\vspace{-12pt}
\end{figure}

\begin{table}[t]
\centering
\caption{\textbf{RoboCasa365 Success Rates.} Values are percentages. ``--'' denotes a suite without a reported result for the corresponding method.}
\vspace{3pt}
\label{tab:main-results}
\small
\resizebox{0.82\textwidth}{!}{%
\begin{tabular}{lcccc}
\toprule
Method & Atomic-Seen & Composite-Seen & Composite-Unseen & Overall \\
\midrule
$\pi_0$ & 34.6 & 6.1 & 1.1 & 14.8 \\
$\pi_{0.5}$ & 39.6 & 7.1 & 1.2 & 16.9 \\
RLDX-1 & 60.0 & 21.3 & 5.0 & 30.0 \\
WorldDreamer & 66.3 & 26.7 & 9.0 & 35.3 \\
Xiaomi-Robotics-1 & 80.2 & 57.1 & 32.1 & 57.4 \\
\midrule
Harness VLA & 91.3 & 55.8 & 40.1 & 63.6 \\
\specialrule{\lightrulewidth}{\aboverulesep}{0pt}
\solidrowcolor{ourslavender}{2.8ex}{0.8ex}
\highlightrowstrut\textbf{EmbodiedRSI (Ours)} & \textbf{94.4} & \textbf{63.1} & \textbf{71.3} & \textbf{77.0} \\
\specialrule{\heavyrulewidth}{0pt}{\belowrulesep}
\end{tabular}}
\vspace{-8pt}
\end{table}

\subsection{Generalization on RoboCasa365 and LIBERO-Pro}
\label{sec:q1}

{\looseness=-1
Composite-Unseen provides the most stringent RoboCasa365 evaluation: familiar skills must support task compositions absent from pretraining. With the robot foundation model frozen, EmbodiedRSI leads the baselines on all three suites and reaches \textbf{77.0\%} overall success (Table~\ref{tab:main-results}). Figure~\ref{fig:robocasa365-tasks} illustrates the three task suites.\par}

On Composite-Unseen, EmbodiedRSI reaches \textbf{71.3\%} success, compared with 40.1\% for the best baseline (Harness VLA). This separation shows where the harness matters most: when previously learned behaviors must be reorganized for a new task.

{\looseness=-1
LIBERO-Pro evaluates generalization under changed instructions and object positions. EmbodiedRSI reaches \textbf{86.8\%} overall success versus 72.1\% for Harness VLA and leads Harness VLA in all eight perturbation cells (Table~\ref{tab:libero-results}). The largest gaps occur under position-swap (S) in SPATIAL and LIBERO-10, indicating that EmbodiedRSI adapts execution when object layouts change. Appendix~\supplementref{app:libero-pro-examples} provides examples of both perturbations. \emph{Together, these results answer \textbf{Q1}: EmbodiedRSI generalizes to unseen task compositions and to changed instructions and object positions.}\par}

\begin{table}[H]
\centering
\caption{\textbf{LIBERO-Pro Success Rates.} The table aggregates success rates (\%) across SPATIAL, OBJECT, GOAL, and LIBERO-10 under instruction-redirection (T) and position-swap (S) perturbations. ``--'' denotes a cell not applicable to the corresponding method.}
\label{tab:libero-results}
\vspace{10pt}
\small
\renewcommand{\arraystretch}{1.12}
\resizebox{\textwidth}{!}{%
\begin{tabular}{lccccccccc}
\toprule
Method & Spat-T & Spat-S & Obj-T & Obj-S & Goal-T & Goal-S & L10-T & L10-S & Overall \\
\midrule
OpenVLA & 0.0 & 0.0 & 0.0 & 0.0 & 0.0 & 0.0 & 0.0 & 0.0 & 0.0 \\
$\pi_0$ & 0.0 & 0.0 & 0.0 & 2.0 & 0.0 & 0.0 & 0.0 & 0.0 & 0.3 \\
$\pi_{0.5}$ & 1.0 & 20.0 & 1.0 & 17.0 & 2.0 & 38.0 & 1.0 & 8.0 & 11.0 \\
MolmoAct & 0.0 & 0.0 & 0.0 & 6.0 & 0.0 & 0.0 & 6.0 & 0.0 & 1.5 \\
NORA & 0.0 & 0.0 & 0.0 & 0.0 & 0.0 & 0.0 & 0.0 & 0.0 & 0.0 \\
X-VLA & 0.0 & 0.0 & 8.0 & 2.0 & 9.0 & 1.0 & 10.0 & 0.0 & 3.8 \\
AtomVLA & 1.0 & 16.0 & 0.0 & 10.0 & 11.0 & 2.0 & 9.0 & 1.0 & 6.3 \\
$\pi_{\text{RLinf}}$ & 42.0 & 59.0 & 71.0 & 78.0 & 45.0 & 42.0 & 49.0 & 14.0 & 50.0 \\
\midrule
CaP-X & 14.0 & 12.0 & 18.0 & 22.0 & 17.0 & 26.0 & -- & -- & 18.2 \\
RATS & 31.0 & 29.0 & 63.0 & 61.0 & 36.0 & 43.0 & -- & -- & 43.8 \\
Harness VLA & 81.0 & 69.0 & 94.0 & 91.0 & 75.0 & 66.0 & 52.0 & 49.0 & 72.1 \\
Zetta & -- & -- & -- & -- & \textbf{92.5} & 89.0 & 63.0 & 40.0 & 71.1 \\
\specialrule{\lightrulewidth}{\aboverulesep}{0pt}
\solidrowcolor{ourslavender}{2.8ex}{0.8ex}
\highlightrowstrut\textbf{EmbodiedRSI (Ours)} & \textbf{100.0} & \textbf{93.0} & \textbf{100.0} & \textbf{96.0} & 78.0 & \textbf{90.0} & \textbf{66.0} & \textbf{71.0} & \textbf{86.8} \\
\specialrule{\heavyrulewidth}{0pt}{\belowrulesep}
\end{tabular}}
\vspace{-10pt}
\end{table}

\subsection{Zero-Shot Sim-to-Real Transfer of the Agentic Harness}
\label{sec:q2}

We next examine whether the gains in simulation carry into physical manipulation by directly deploying EmbodiedRSI's agentic harness, including code and skills evolved in simulation, on a physical SO-101 arm~\citep{cadene2026lerobot}. SmolVLA~\citep{shukor2025smolvla} serves as the robot foundation model. We compare SmolVLA alone with the same backbone under EmbodiedRSI's agentic harness. For each task, we collect 50 human-teleoperated demonstrations and fully fine-tune SmolVLA for 20,000 steps on 4$\times$ NVIDIA H200 GPUs. The model remains frozen during all trials. We run 30 trials per condition on each task. The remaining training settings are given in Appendix~\supplementref{app:real-robot-training}.

These five highly challenging real-world tasks span three categories: \emph{(1) Multi-Step Manipulation (cake stacking and towel folding), (2) Semantic and Arithmetic Reasoning (Oreo placement and poker-chip selection), and (3) Precision Grasping (lifting glasses at the bridge).}

\vspace{14pt}
\noindent
\begin{minipage}[t]{0.43\textwidth}
\vspace{0pt}
The tasks require multi-step execution, grounding object descriptions in the scene, and precise control at a narrow grasp point.

In the poker-chip task, the robot must push out the chip whose value multiplied by five equals 100. The target is the chip marked 20 because $20\times5=100$. We shuffle the chips' order and orientation before every trial. Figure~\ref{fig:real-robot-tasks} illustrates the first three tasks. Table~\ref{tab:real-robot-results} reports success on all five. The largest gain occurs on poker-chip selection, where the agent pipeline identifies the correct chip from the arithmetic instruction before directing SmolVLA in a randomized layout.

\noindent\emph{With SmolVLA frozen, zero-shot transfer of EmbodiedRSI's agentic harness raises overall real-robot success from 46.0\% to 71.3\%, answering \textbf{Q2}: improvements learned by the harness in simulation carry over to physical manipulation.}
\end{minipage}\hfill
\begin{minipage}[t]{0.55\textwidth}
\vspace{0pt}\vspace{-26.04pt}
\centering
\makebox[\linewidth][r]{\includegraphics[width=0.95\linewidth]{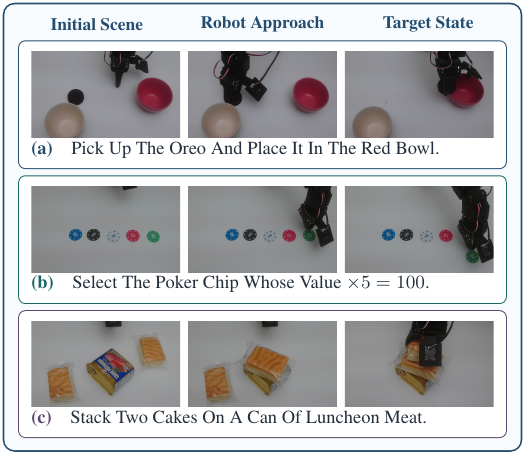}}
\setlength{\abovecaptionskip}{4pt}
\captionof{figure}{\textbf{Representative Real-World Evaluation Tasks.} Initial and target states.}
\label{fig:real-robot-tasks}
\end{minipage}
\par

\begin{table}[H]
\centering
\caption{\textbf{Real-Robot Success Rates.} Success over 30 trials per task on the real robot.}
\vspace{3pt}
\label{tab:real-robot-results}
\small
\resizebox{0.68\textwidth}{!}{
\begin{tabular}{lcc}
\toprule
Task & SmolVLA & \textbf{EmbodiedRSI} \\
\midrule
Oreo to Red Bowl & 70.0\% & 86.7\% \\
Poker Chip Selection ($\times5=100$) & 10.0\% & 90.0\% \\
Stack Two Cakes on a Can of Luncheon Meat & 63.3\% & 76.7\% \\
Glasses Bridge Grasp & 60.0\% & 56.7\% \\
Towel Folding & 26.7\% & 46.7\% \\
\specialrule{\lightrulewidth}{\aboverulesep}{0pt}
\solidrowcolor{ourslavender}{2.8ex}{0.8ex}
\highlightrowstrut\textbf{Overall} & \textbf{46.0\%} & \textbf{71.3\%} \\
\specialrule{\heavyrulewidth}{0pt}{\belowrulesep}
\end{tabular}}
\vspace{-8pt}
\end{table}


\subsection{Ablation Study}
\label{sec:ablation}

We further explore what drives EmbodiedRSI's gains from physical interaction (\textbf{Q3}). The ablations isolate how Code-Skill Co-Evolution, Value-of-Information Experiment Selection, and Reward-Grounded Memory Learning affect success and learning efficiency. For the code-skill comparison, we hold the frozen VLA, task suites, physical-experiment budget, and evaluation protocol fixed. Code-only Evolution updates executable code while keeping skills fixed, Skill-only Evolution updates skills while keeping executable code fixed, and Code-Skill Co-Evolution updates both code and skills.

\begin{figure}[H]
\noindent
\begin{minipage}[h]{0.44\textwidth}
\raggedright
Figure~\ref{fig:code-skill-ablation} shows that Code-Skill Co-Evolution improves success across RoboCasa365, with the clearest advantage on Composite-Unseen tasks. On this suite, Code-Skill Co-Evolution reaches 71.3\% success versus 55.6\% for Skill-only Evolution. Notably, Code-only Evolution outperforms Skill-only Evolution on Atomic-Seen, but the order reverses on Composite-Unseen. This reversal suggests that a change effective for familiar atomic tasks may not carry over to unseen compositions. Joint evolution leads on both suites, indicating that code and skills are more useful when updated together.
\end{minipage}\hfill
\begin{minipage}[h]{0.54\textwidth}
\centering
\vspace{-5pt}
\includegraphics[width=\linewidth]{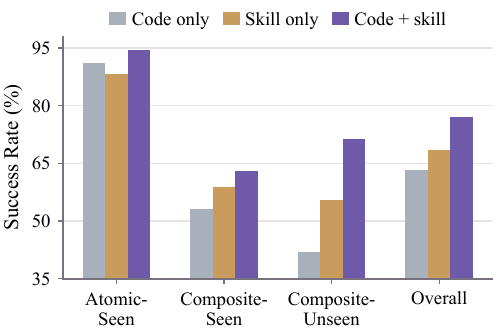}
\vspace{-12pt}
\caption{\textbf{Code-Skill Co-Evolution Ablation.} Bars show RoboCasa365 success rates.}
\label{fig:code-skill-ablation}
\vspace{0pt}
\end{minipage}
\end{figure}

\begin{figure}[H]
\centering
\includegraphics[width=0.88\textwidth]{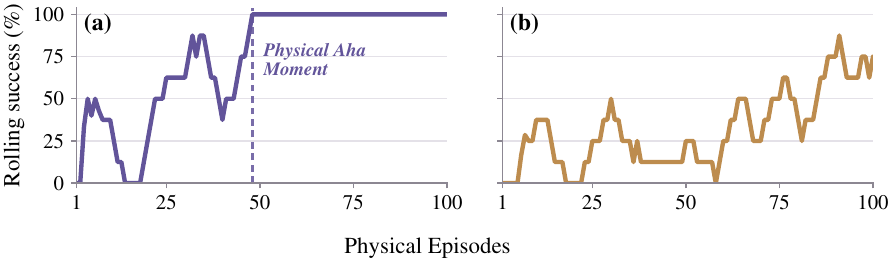}
\vspace{-8pt}
\caption{\textbf{Representative LIBERO-Pro Learning Curves.} Eight-episode rolling success on SPAT-T over 100 physical episodes. \textbf{(a)} Value-of-Information Experiment Selection reaches sustained 100\% at episode 48 (dashed line). \textbf{(b)} Random Selection continues to fluctuate.}
\label{fig:voi-ablation}
\vspace{-8pt}
\end{figure}

\textbf{Learning Efficiency.} Table~\ref{tab:learning-efficiency} compares EmbodiedRSI with Random Exploration after 50 physical episodes. EmbodiedRSI reaches 72.9\% overall success versus 27.1\%, and its learning AUC is 0.635 versus 0.309. A \textbf{\emph{Physical Aha Moment}} for agentic robotics occurs when an agent identifies the critical physical state variable needed to return a VLA to its in-domain region~\citep{ding2026zetta}. In a separate representative LIBERO-Pro SPAT-T case, Figure~\ref{fig:voi-ablation} shows the observable marker of this moment: Value-of-Information Experiment Selection reaches 100\% eight-episode rolling success within 50 physical episodes and remains there. Random Selection of code and skill experiments continues to fluctuate through episode 100 and ends around 75.0\% rolling success, suggesting that more physical episodes are needed for a comparable \textbf{\emph{Physical Aha Moment}}.

\begin{table}[H]
\centering
\caption{\textbf{Learning Efficiency.} EmbodiedRSI is compared with Random Exploration.}
\vspace{12pt}
\label{tab:learning-efficiency}
\fontsize{8.5pt}{10pt}\selectfont
\renewcommand{\arraystretch}{1.2}
\begin{tabular}{lcc}
\toprule
Method & Overall Success @ 50 Ep. $\uparrow$ & Learning AUC $\uparrow$ \\
\midrule
Random Exploration & 27.1 & 0.309 \\
\specialrule{\lightrulewidth}{\aboverulesep}{0pt}
\solidrowcolor{ourslavender}{2.8ex}{0.8ex}
\highlightrowstrut\textbf{EmbodiedRSI (Ours)} & \textbf{72.9} & \textbf{0.635} \\
\specialrule{\heavyrulewidth}{0pt}{\belowrulesep}
\end{tabular}
\vspace{-7pt}
\end{table}

\textbf{Reward-Grounded Memory Learning.} Table~\ref{tab:rl-ablation} compares retrieval-based selection, PPO with task reward alone, and the full method. The ablation settings are detailed in \supplementautoref{app:memory}. Composite-Unseen success rises from 61.7\% and 65.2\% to 71.3\%. Learning AUC rises from 0.556 and 0.589 to 0.635. \emph{Together, the ablations resolve \textbf{Q3}: Code-Skill Co-Evolution, hypothesis-guided exploration, and Reward-Grounded Memory Learning improve success and learning efficiency.}

\begin{table}[H]
\centering
\caption{\textbf{Reward-Grounded Memory Learning Ablation.} Composite-Unseen and learning AUC on RoboCasa365.}
\vspace{5pt}
\label{tab:rl-ablation}
\fontsize{8.5pt}{10pt}\selectfont
\renewcommand{\arraystretch}{0.9}
\begin{tabular}{lcc}
\toprule
Variant & Composite-Unseen $\uparrow$ & Learning AUC $\uparrow$ \\
\midrule
Retrieval-Based & 61.7\% & 0.556 \\
PPO + Task Reward Only & 65.2\% & 0.589 \\
\specialrule{\lightrulewidth}{\aboverulesep}{0pt}
\solidrowcolor{ourslavender}{1.85ex}{0.55ex}
\rule[-0.55ex]{0pt}{2.4ex}\textbf{EmbodiedRSI (Ours)} & \textbf{71.3\%} & \textbf{0.635} \\
\specialrule{\heavyrulewidth}{0pt}{\belowrulesep}
\end{tabular}
\vspace{-7pt}
\end{table}

\section{Conclusion}
EmbodiedRSI shows that a self-evolving agentic harness can gain lasting value from each physical trial by using the outcome to select the next experiment, refine code and skills together, and retain experience for later adaptation. Across simulated benchmarks and real-robot tasks, EmbodiedRSI improves generalization and learning efficiency, and transfers the simulation-evolved harness to physical execution. More broadly, continual robot learning can take place in the harness around a frozen robot foundation model when physical evidence shapes both the next experiment and the knowledge carried into future tasks.

\noindent\textbf{Limitations.} Code-as-policy and agentic harness methods learn executable behavior from physical trajectories without updating the robot foundation model's parameters. Future work could use trajectories selected by the harness for online model updates, allowing code, skills, and the model to improve together.

\bibliography{references}
\bibliographystyle{bibstyle}

\ifincludeappendix
\clearpage
\appendix
\setcounter{figure}{0}
\renewcommand{\thefigure}{A.\arabic{figure}}
\renewcommand{\theHfigure}{appendix.\arabic{figure}}
\setcounter{table}{0}
\counterwithin{table}{section}
\renewcommand{\thetable}{\thesection.\arabic{table}}
\renewcommand{\theHtable}{appendix.\thesection.\arabic{table}}
\section*{Appendix}
\label{app:appendix}








\section{LIBERO-Pro Perturbation Examples}
\label{app:libero-pro-examples}
\begin{figure}[h]
\centering
\begin{minipage}[t]{0.48\textwidth}
    \centering
    \begin{minipage}[t]{0.45\linewidth}
        \centering
        {\setlength{\fboxsep}{1.5pt}\colorbox{positiontint}{\makebox[\dimexpr\linewidth-2\fboxsep\relax][c]{\tiny\bfseries Original Layout}}}
        \par\vspace{1pt}
        \includegraphics[width=\linewidth]{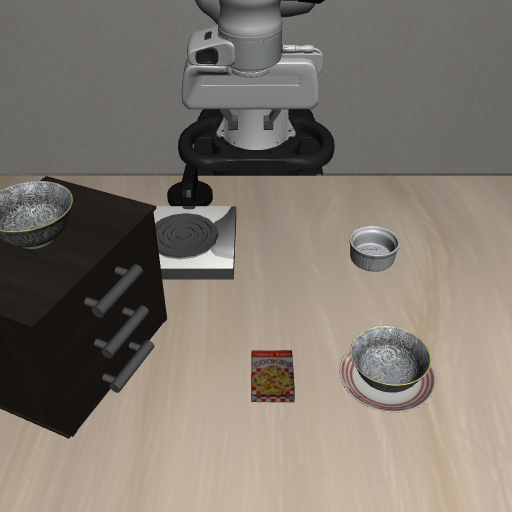}
    \end{minipage}
    \hfill\raisebox{-0.21\linewidth}{\(\rightarrow\)}\hfill
    \begin{minipage}[t]{0.45\linewidth}
        \centering
        {\setlength{\fboxsep}{1.5pt}\colorbox{positiontint}{\makebox[\dimexpr\linewidth-2\fboxsep\relax][c]{\tiny\bfseries Object Shift}}}
        \par\vspace{1pt}
        \includegraphics[width=\linewidth]{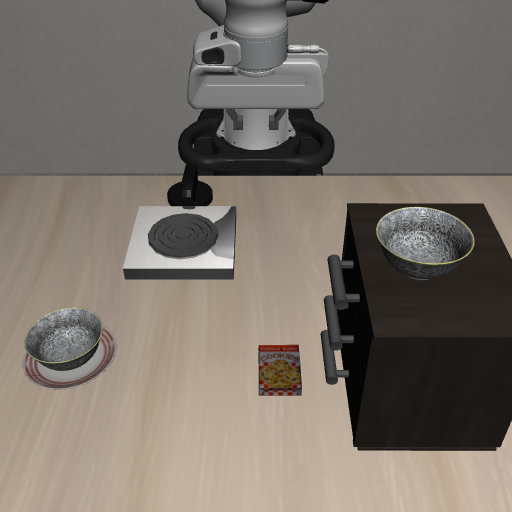}
    \end{minipage}
    \par\smallskip\footnotesize\textbf{(a)} Position Perturbation
\end{minipage}\hfill
\begin{minipage}[t]{0.48\textwidth}
    \centering
    \begin{minipage}[t]{0.45\linewidth}
        \centering
        {\setlength{\fboxsep}{1.5pt}\colorbox{tasktint}{\makebox[\dimexpr\linewidth-2\fboxsep\relax][c]{\tiny\bfseries One Object}}}
        \par\vspace{1pt}
        \includegraphics[width=\linewidth]{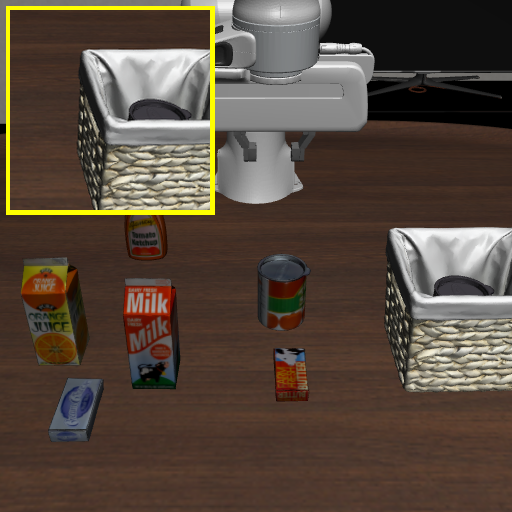}
    \end{minipage}
    \hfill\raisebox{-0.21\linewidth}{\(\rightarrow\)}\hfill
    \begin{minipage}[t]{0.45\linewidth}
        \centering
        {\setlength{\fboxsep}{1.5pt}\colorbox{tasktint}{\makebox[\dimexpr\linewidth-2\fboxsep\relax][c]{\tiny\bfseries Two Objects}}}
        \par\vspace{1pt}
        \includegraphics[width=\linewidth]{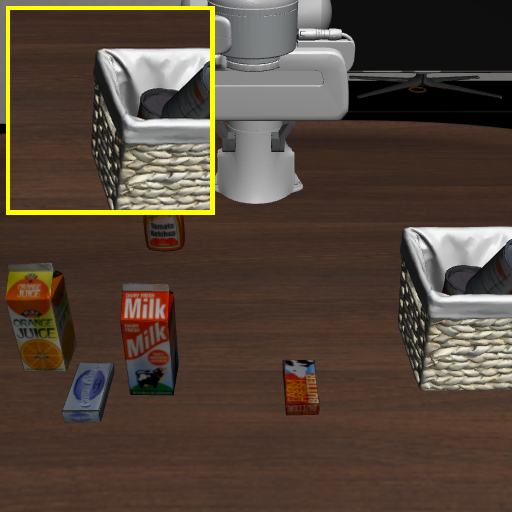}
    \end{minipage}
    \par\smallskip\footnotesize\textbf{(b)} Task Perturbation
\end{minipage}
\caption{\textbf{Representative LIBERO-Pro Perturbation Settings.} \textbf{(a) Position Perturbation} relocates manipulable objects within feasible spatial bounds, illustrated by moving the cabinet, bowls, and cookie box. \textbf{(b) Task Perturbation} redefines task logic and target states: the depicted task requires two objects in the goal basket rather than one. Yellow boxes indicate the goal basket~\citep{zhou2025liberopro}.}
\label{fig:libero_pro_generalization}
\end{figure}

\section{Real-Robot Training Configuration}
\label{app:real-robot-training}
Table~\ref{tab:real-robot-training} lists the complete SmolVLA fine-tuning configuration used for the real-robot comparison.

\begin{table}[h]
\centering
\caption{\textbf{SmolVLA Fine-Tuning Configuration.} Settings shared by both real-robot conditions before the agent pipeline is applied.}
\label{tab:real-robot-training}
\small
\begin{tabular}{ll}
\toprule
Parameter & Setting \\
\midrule
Hardware & 4$\times$ NVIDIA H200 \\
Parallelism & Distributed data parallel (\texttt{--multi\_gpu --num\_processes=4}) \\
Precision & bfloat16 \\
Fine-tuning scope & Full model \\
Freeze vision encoder & \texttt{freeze\_vision\_encoder=false} \\
Train expert only & \texttt{train\_expert\_only=false} \\
Batch size per GPU & 64 \\
Effective batch size & 256 \\
Training steps & 20,000 \\
Learning-rate schedule & Cosine decay \\
Warmup steps & 1,000 \\
Decay steps & 20,000 \\
\bottomrule
\end{tabular}
\end{table}

The training loss decreases from 0.65 to 0.013, the gradient norm stabilizes near 0.2, and wall-clock training time is approximately 4 hours 7 minutes. Both real-robot conditions use this backbone. EmbodiedRSI adds its agent pipeline during evaluation.

\section{Hierarchical Memory Schema}
\label{app:memory}
EmbodiedRSI records an episode as a sequence of physical transitions. Hierarchical Memory stores each raw transition as M1 experience, organizes executable code and skill hypotheses as M2, and periodically extracts reusable cognition as M3. The M1 record contains tool arguments, a pre-state, a post-state, execution latency, raw tool output, a physical-outcome label, and task reward. A state can contain end-effector pose, gripper state, task progress, VLA history, and simulator-only fields. The planner accesses tool-interface fields, while simulator-only fields remain in the memory record. The resulting hierarchy links episodes, physical states, hypotheses, outcomes, and content-addressed artifacts. Table~\ref{tab:memory-roles} summarizes the component responsibilities, and Table~\ref{tab:memory-actions} defines the Slow-System action space.

\begingroup
\setlength{\intextsep}{6pt}
\begin{table}[H]
\centering
\caption{\textbf{Hierarchical Memory Responsibilities.} Components, inputs, and responsibilities in the Fast-Slow loop.}
\label{tab:memory-roles}
\small
\setlength{\tabcolsep}{5pt}
\renewcommand{\arraystretch}{1.15}
\begin{tabular}{@{}p{0.17\textwidth}p{0.26\textwidth}p{0.52\textwidth}@{}}
\toprule
Component & Input & Responsibility \\
\midrule
Manager & Current experience and retrieved M1, M2, M3 records & Select an applicable action and target using the PPO-trained policy. \\
Selector & Current experience and indexed records & Retrieve relevant M1, M2, and M3 records with evidence links. \\
Operator & Selected action, target, and source records & Generate and apply a structured update using a frozen LLM. \\
KnowledgeStore & Experience, hypothesis states, and cognition & Store M1--M3 records, evidence links, and retrieval indices. \\
M1: Raw Experience & Execution trajectory and outcome & Automatically append every trial record and preserve the raw evidence. \\
M2: Hypothesis State & Matched M1 evidence & Maintain hypothesis counts and graph relations under the Fast-System update rules. \\
M3: Reusable Cognition & Selected action and source evidence & Store reusable knowledge with conditions, strategies, scope, and evidence references. \\
\bottomrule
\end{tabular}
\par\vspace{8pt}
\normalsize
\caption{\textbf{Slow-System Memory Actions.} Each action specifies how experience contributes to reusable knowledge.}
\label{tab:memory-actions}
\small
\setlength{\tabcolsep}{5pt}
\renewcommand{\arraystretch}{1.15}
\begin{tabular}{@{}p{0.14\textwidth}p{0.55\textwidth}p{0.26\textwidth}@{}}
\toprule
Action & Purpose & Target \\
\midrule
Retain & Add valuable experience to the long-term retrieval index. & M1 retrieval index \\
Merge & Incorporate new evidence into existing cognition while preserving applicable conditions. & Existing M3 item \\
Abstract & Create new reusable cognition from recurring experience and hypothesis evidence. & M1 and M2 evidence group $\rightarrow$ M3 \\
Compress & Shorten the representation while preserving meaning and evidence links. & Indexed representation or M3 item \\
Forget & Remove an outdated or superseded entry from active retrieval while preserving source evidence. & Retrieval index or active M3 entry \\
Skip & Leave the retrieval index and M3 unchanged. & No update \\
\bottomrule
\end{tabular}
\end{table}

Each candidate $s_i$ identifies an applicable action and target record or evidence group. The manager scores candidates using the current experience and retrieved records, including task context, hypothesis confidence, and evidence links. Skip is always available. The operator applies the selected action while preserving raw M1 records and evidence-driven M2 updates. After the operator applies a memory action, the Fast System selects and executes the next experiment using the updated memory. The resulting success-rate change, entropy reduction, and realized experiment cost provide the reward. The manager is trained with PPO on discounted returns over $L$ memory decisions. Here $H_t$ denotes $\mathcal{H}_{\mathrm{exp}}$ at decision $t$, with both entropy estimates evaluated over the same hypotheses. $C_{t+1}$ uses the elapsed time, simulator steps, GPU cost, and reset cost in $C(\cdot)$.

\endgroup

\myparagraph{Controlled Memory Ablation Protocol.}
Table~\ref{tab:memory-ablation-protocol} documents the three memory-selection variants used in Table~\ref{tab:rl-ablation}. This comparison separates learned action selection from retrieval-based selection and evaluates the joint contribution of entropy reduction and physical-experiment cost in the reward.

\begin{table}[htbp]
\centering
\caption{\textbf{Controlled Memory Ablation Protocol.} Selection rules and training rewards used in the three ablation variants.}
\label{tab:memory-ablation-protocol}
\small
\setlength{\tabcolsep}{5pt}
\renewcommand{\arraystretch}{1.15}
\begin{tabular}{@{}p{0.23\textwidth}p{0.40\textwidth}p{0.32\textwidth}@{}}
\toprule
Variant & Selection Rule & Training Reward \\
\midrule
Retrieval-Based & Select the applicable action and target with the highest cosine similarity between the current experience and candidate description under the frozen encoder. & No PPO training \\
PPO + Task Reward Only & PPO Manager selects an applicable action and target. & $\lambda_1(R_{t+1}-R_t)$ \\
EmbodiedRSI & The same PPO Manager selects an applicable action and target. & $\begin{aligned}[t]r_t^{\mathrm{mem}}={}&\lambda_1(R_{t+1}-R_t)\\ &+\lambda_2(H_t-H_{t+1})\\ &-\lambda_3 C_{t+1}\end{aligned}$ \\
\bottomrule
\end{tabular}
\end{table}

Across these experiments, we hold the remaining components constant. Each Manager receives the current experience and retrieved M1, M2, and M3 records, including hypothesis confidence and evidence links. All variants share the six-action space, target construction, applicability rules, and Skip availability. The encoder, frozen Qwen operator, operator prompts, and generation settings are identical. Action instructions remain unchanged during this comparison. Fast-System hypothesis updates, Code-Skill Co-Evolution, and Value-of-Information Experiment Selection follow the same rules. All variants start from the same memory and hypotheses and receive equal physical-interaction budgets, task distributions, and evaluation protocols. Only the selection rule or training reward differs.

\section{Value-of-Information Experiment Selection}
\label{app:voi}
\noindent\textbf{Notation.} Let $\mathcal{G}$ be the Hypothesis Graph, $\mathcal{H}$ its code and skill hypotheses, and $D_k$ the matched physical outcomes observed after $k$ episodes. An experiment $t$ may invoke any ordered sequence of code and skill nodes. Its tested set $S_t\subseteq\mathcal{H}$ identifies the nodes evaluated by matched comparisons. For each tested node, the comparison preserves the initial state and all other calls in their original order. A positive success-rate difference adds one to $n_{\mathrm{sup},i}$, a negative difference adds one to $n_{\mathrm{ref},i}$, and a tie adds one to $n_{\mathrm{tie},i}$. The evaluation count is $n_{\mathrm{eval},i}=n_{\mathrm{sup},i}+n_{\mathrm{ref},i}+n_{\mathrm{tie},i}$. Each tested node is compared once per experiment. Comparisons of several tested nodes use independent resets, and their physical episodes are included in $q(t)$.

For hypothesis $h_i$, let $\theta_i$ be the probability of a positive difference when the matched outcomes differ. A shared $\operatorname{Beta}(\alpha,\beta)$ prior yields $\theta_i\mid D_k\sim\operatorname{Beta}(a_i,b_i)$, where $a_i=\alpha+n_{\mathrm{sup},i}$ and $b_i=\beta+n_{\mathrm{ref},i}$. Let $\rho_i$ be the probability that the matched outcomes differ, with shared prior $\operatorname{Beta}(\gamma,\delta)$ and posterior parameters $u_i=\gamma+n_{\mathrm{sup},i}+n_{\mathrm{ref},i}$ and $v_i=\delta+n_{\mathrm{tie},i}$. The probabilities for the next matched comparison are:
\begin{equation}
\begin{aligned}
P(e_i=+\mid D_k)&=\frac{u_i}{u_i+v_i}\frac{a_i}{a_i+b_i}, &
P(e_i=-\mid D_k)&=\frac{u_i}{u_i+v_i}\frac{b_i}{a_i+b_i},\\
P(e_i=0\mid D_k)&=\frac{v_i}{u_i+v_i}.
\end{aligned}
\end{equation}
For several tested nodes, $P(e\mid D_k,t)$ is the product of these probabilities across their independently reset comparisons. Both entropy terms in $G_k(t)$ use the current hypothesis set. New hypotheses enter the next round with the same priors and receive an initial matched comparison before $G_k(t)$ is used to select their next experiment. The budget is $B$ physical episodes, $n_{\max}$ is the evaluation cap per hypothesis, and $\epsilon_H,\epsilon_{\mathrm{VOI}}$ are the stopping thresholds. $\operatorname{Candidates}$ returns experiments with nonempty tested sets that fit the remaining budget and cap. $\operatorname{UpdateGraph}$ updates the three outcome counts and supported $\textsc{works-with}$ edges. $\operatorname{Propose}$ returns new code or skill hypotheses and revisions from the recorded evidence. $\operatorname{Insert}$ gives every proposal the shared priors and adds a $\textsc{refines}$ edge only for a revision.

\begingroup
\captionof{algorithm}{Hypothesis Graph Confidence and Value-of-Information Experiment Selection}
\label{alg:hypothesis-voi}
\small
\begin{algorithmic}[1]
\Require Initial graph $\mathcal{G}$; shared priors $\alpha,\beta,\gamma,\delta$; budget $B$; cap $n_{\max}$; thresholds $\epsilon_H,\epsilon_{\mathrm{VOI}}$
\State $D,\mathcal{M}_1\gets\emptyset$; $b\gets0$; $n_{\mathrm{sup},i},n_{\mathrm{ref},i},n_{\mathrm{tie},i}\gets0$ for all initial $h_i$
\While{$b<B$}
    \State $\mathcal{H}\gets\operatorname{nodes}(\mathcal{G})$; $\mathcal{U}\gets\{h_i\in\mathcal{H}:n_{\mathrm{eval},i}<n_{\max}\}$
    \State $\mathcal{T}\gets\operatorname{Candidates}(\mathcal{G},\mathcal{U},B-b,n_{\max})$
    \If{$\mathcal{T}=\emptyset$} \State \textbf{break} \EndIf
    \State $\mathcal{T}_0\gets\{t\in\mathcal{T}:S_t\cap\{h_i:n_{\mathrm{eval},i}=0\}\neq\emptyset\}$
    \If{$\mathcal{T}_0\neq\emptyset$}
        \State $t^*\sim\operatorname{Uniform}(\mathcal{T}_0)$
    \Else
        \State $q_i\gets P(\theta_i>1/2\mid D)$; $\mathcal{H}_{\mathrm{exp}}(D)\gets\sum_{h_i\in\mathcal{H}}h_2(q_i)$
        \For{each $t\in\mathcal{T}$}
            \State $G_b(t)\gets\mathcal{H}_{\mathrm{exp}}(D)-\mathbb{E}_{e\mid D,t}[\mathcal{H}_{\mathrm{exp}}(D\cup\{e\})]$
            \State $\operatorname{VOI}_b(t)\gets G_b(t)/C(t)$
        \EndFor
        \If{$\mathcal{H}_{\mathrm{exp}}(D)<\epsilon_H$ or $\max_{t\in\mathcal{T}}\operatorname{VOI}_b(t)<\epsilon_{\mathrm{VOI}}$}
            \State \textbf{break}
        \EndIf
        \State $t^*\gets\arg\max_{t\in\mathcal{T}}\operatorname{VOI}_b(t)$
    \EndIf
    \State $e\gets\operatorname{Execute}(t^*)$; $b\gets b+q(t^*)$; $D\gets D\cup\{e\}$; $\mathcal{M}_1\gets\mathcal{M}_1\cup\{e\}$
    \State $\mathcal{G}\gets\operatorname{UpdateGraph}(\mathcal{G},t^*,e)$
    \State $(\mathcal{P}_{\mathrm{new}},\mathcal{P}_{\mathrm{rev}})\gets\operatorname{Propose}(\mathcal{G},\mathcal{M}_1,e)$
    \State $\mathcal{G}\gets\operatorname{Insert}(\mathcal{G},\mathcal{P}_{\mathrm{new}},\mathcal{P}_{\mathrm{rev}})$
\EndWhile
\State $c_i\gets a_i/(a_i+b_i)$; $q_i\gets P(\theta_i>1/2\mid D)$ for all $h_i\in\operatorname{nodes}(\mathcal{G})$
\State \Return $\mathcal{G},\{c_i,q_i\}$
\end{algorithmic}
\endgroup

\section{Fast-System Experimental Configuration}
\label{app:fast-system-config}
Table~\ref{tab:fast-reference-config} specifies the configuration used in our experiments. The cost weights express time in seconds, simulator steps as counts, GPU use in GPU-seconds, and reset cost as the number of resets. The entropy threshold scales with the current hypothesis count, corresponding to an average posterior entropy below $0.1$ bit per hypothesis.

\begin{table}[H]
\centering
\caption{\textbf{Fast-System Experimental Configuration.} Numerical settings and online evaluation budget used in our experiments.}
\label{tab:fast-reference-config}
\small
\setlength{\tabcolsep}{5pt}
\renewcommand{\arraystretch}{1.12}
\begin{tabular}{@{}p{0.19\textwidth}p{0.29\textwidth}p{0.47\textwidth}@{}}
\toprule
Parameter & Value & Definition \\
\midrule
$\alpha,\beta$ & $1,1$ & Beta prior for hypothesis support \\
$\gamma,\delta$ & $1,1$ & Beta prior for matched-outcome disagreement \\
$w_\tau$ & $0.01$ & Weight on elapsed execution time \\
$w_s$ & $2.5\times10^{-4}$ & Weight per simulator step \\
$w_g$ & $0.005$ & Weight on GPU-seconds \\
$w_r$ & $0.10$ & Weight per environment reset \\
$\epsilon_H$ & $0.1|\mathcal H|$ bits & Threshold for total hypothesis entropy \\
$\epsilon_{\mathrm{VOI}}$ & $10^{-3}$ & Minimum VOI for continuing exploration \\
$n_{\max}$ & $6$ & Maximum matched evaluations per hypothesis \\
$B$ & $200$ episodes per task run & Maximum physical-interaction budget \\
Reporting window & Final 100 episodes & Online success while the harness evolves \\
\bottomrule
\end{tabular}
\end{table}

The online metric averages the binary task outcomes of the final 100 physical episodes. Each outcome evaluates the harness available at that episode. Reaching a stopping criterion ends broad hypothesis exploration, while task execution continues to complete the reporting window. We expect additional iterations to offer further opportunities for improving task success rate.

\section{PPO Training Configuration}
\label{app:ppo-training}
The Manager is trained with PPO during evolution and remains frozen during evaluation. Table~\ref{tab:ppo-training} lists the settings used to train the Manager. Each rollout collects $L=32$ memory decisions before four PPO optimization epochs. The encoder and operator remain frozen.

\begin{table}[H]
\centering
\caption{\textbf{PPO Training Configuration.} Hyperparameters used for Manager training.}
\label{tab:ppo-training}
\small
\setlength{\tabcolsep}{5pt}
\renewcommand{\arraystretch}{1.12}
\begin{tabular}{@{}p{0.32\textwidth}p{0.20\textwidth}p{0.43\textwidth}@{}}
\toprule
Parameter & Value & Definition \\
\midrule
Optimizer & AdamW & Policy and value optimization \\
Learning rate & $2.5\times10^{-4}$ & Policy and value learning rate \\
PPO clip ratio & $0.18$ & Clipping range for the policy objective \\
Rollout horizon $L$ & $32$ & Memory decisions per PPO rollout \\
PPO epochs & $4$ & Optimization passes per rollout \\
Mini-batch size & $16$ & Samples per gradient step \\
Discount $\gamma$ & $0.97$ & Discount for future memory rewards \\
GAE $\lambda$ & $0.92$ & Advantage-estimation smoothing \\
Entropy coefficient & $0.012$ & Weight on policy entropy \\
Value-loss coefficient & $0.5$ & Weight on value-function loss \\
Maximum gradient norm & $0.8$ & Gradient-clipping threshold \\
Training seeds & $10$ & Number of independent training seeds \\
\bottomrule
\end{tabular}
\end{table}

\section{Hypothesis Graph Relations and Cost Model}
\label{app:hgraph}
The Hypothesis Graph contains executable code and skill hypotheses as nodes. A directed $\textsc{refines}$ edge records version evolution from a parent hypothesis to a more specific hypothesis. A bidirectional $\textsc{works-with}$ edge records measured positive joint gain between a code hypothesis and a skill hypothesis. The Hypothesis Graph keeps significant $\textsc{works-with}$ edges with the highest joint gain and retains $\textsc{refines}$ edges as version history. Candidate physical experiments declare the hypothesis subsets they distinguish and are ranked by the cost model in \autoref{sec:hgraph}. The cost model accounts for elapsed time, simulator steps, GPU use, and reset cost, so the Fast System searches lower-cost experiments before more expensive physical trials.

\section{Trials from Common Initial States}
\label{app:coevo}
For each code hypothesis, skill hypothesis, or code-skill pair, EmbodiedRSI runs trials from the same initial states. The resulting success rate defines $R(\cdot)$ in the $\textsc{works-with}$ score. Every trial outcome is written to M1, updates the corresponding M2 hypothesis confidence, and can contribute to M3 cognition after repeated evidence supports a reusable pattern.

\section{Shared Backbones for Baseline Comparisons}
\label{app:shared-backbones}
Table~\ref{tab:shared-backbones} specifies the shared robot foundation models used in comparisons between agentic harnesses. All compared harnesses use identical backbone weights and evaluation settings on each benchmark. Only the harness differs.

\begin{table}[htbp]
\centering
\caption{\textbf{Shared Backbones for Baseline Comparisons.} Backbone settings shared across the compared harnesses.}
\label{tab:shared-backbones}
\small
\setlength{\tabcolsep}{5pt}
\renewcommand{\arraystretch}{1.15}
\begin{tabular}{@{}p{0.18\textwidth}p{0.50\textwidth}p{0.27\textwidth}@{}}
\toprule
Benchmark & Shared Backbone and Weights & Difference \\
\midrule
RoboCasa365 & Xiaomi-Robotics-1, identical weights & Agentic harness \\
LIBERO-Pro & $\pi_{0.5}$, official RLinf fine-tuned weights & Agentic harness \\
\bottomrule
\end{tabular}
\end{table}

\fi

\end{document}